\documentclass[final]{siamltex}
\usepackage{tikz}
\usepackage{amsfonts}
\usepackage{epsfig}
\usepackage{color}
\usepackage{amsmath}
\usepackage{bm}
\usepackage{booktabs}

\begin{document} 

\title{Deep Multi-fidelity Gaussian Processes}
\date{}
\author{\bf{Maziar Raissi \& George Karniadakis}}
\maketitle

\begin{abstract}
We develop a novel multi-fidelity framework that goes far beyond the classical AR(1) Co-kriging scheme of Kennedy and O'Hagan (2000). Our method can handle general discontinuous cross-correlations among systems with different levels of fidelity. A combination of multi-fidelity Gaussian Processes (AR(1) Co-kriging) and deep neural networks enables us to construct a method that is immune to discontinuities. We demonstrate the effectiveness of the new technology using standard benchmark problems designed to resemble the outputs of complicated high- and low-fidelity codes. 
\end{abstract}


\section{Motivation} Multi-fidelity modeling proves extremely useful while solving inverse problems for instance. Inverse problems are ubiquitous in science. In general, the response of a system is modeled as a function $y = g(x)$. The goal of model inversion is to find a parameter setting $x$ that matches a target response $y^* = g(x^*)$. In other words, we are solving the following optimization problem:
\[
\min_{\bm{x}}||g(\bm{x}) - y^*||,
\]
for some suitable norm. In practice, $\bm{x}$ is often a high-dimensional vector and $g$ is a complex, non-linear, and expensive to compute map. These factors render the solution of the optimization problem very challenging and motivate the use of surrogate models as a remedy for obtaining inexpensive samples of $g$ at unobserved locations. To this end, a surrogate model acts as an intermediate agent that is trained on available realizations of $g$, and then is able to perform accurate predictions for the response at a new set of inputs. Multi-fidelity framework can be employed to build efficient surrogate models of $g$. Our Deep Multi-fidelity GP algorithm is most useful when the function $g$ is very complicated, involves discontinuities, and when the correlation structures between different levels of fidelity have discontinuous nonfunctional forms.

\section{Introduction} Using deep neural networks, we build a multi-fidelity model that is immune to discontinuities. We employ Gaussian Processes (GPs) (see \cite{williams2006gaussian}) which is a non-parametric Bayesian regression technique. Gaussian Processes is a very popular and useful tool to approximate an objective function given some of its observations. It corresponds to a particular class of surrogate models which makes the assumption that the response of the complex system is a realization of a Gaussian process. In particular, we are interested in Manifold Gaussian Processes \cite{calandra2014manifold} that are capable of capturing discontinuities. Manifold GP is equivalent to jointly learning a data transformation into a feature space followed by a regular GP regression. The model profits from standard GP properties. We show that the well-known classical multi-fidelity Gaussian Processes (AR(1) Co-kriging) \cite{kennedy2000predicting} is a special case of our method. Multi-fidelity modeling is most useful when low-fidelity versions of a complex system are available. They may be less accurate but are computationally cheaper.
\\

For the sake of clarity of presentation, we focus only on two levels of fidelity. However, our method can be readily generalized to multiple levels of fidelity. In the following, we assume that we have access to data with two levels of fidelity $\left\{ \{\bm{x}_1, \bm{f}_1\}, \{\bm{x}_2,\bm{f}_2\}\right\}$, where $\bm{f}_2$ has a higher level of fidelity. We use $n_1$ to denote the number of observations in $\bm{x}_1$ and $n_2$ to denote the sample size of $\bm{x}_2$. The main assumption is that $n_2 < n_1$. This is to reflect the fact that high-fidelity data are scarce since they are generated by an accurate but costly process. The low fidelity data, on the other hand, are less accurate, cheap to generate and hence are abundant.\\

As for the notation, we employ the following convention. A boldface letter such as $\bm{x}$ is used to denote data. A non-boldface letter such as $h$ is used to denote both a vector or a scalar. This will be clear from the context.

\section{Deep Multi-fidelity Gaussian Processes} A simple way to explain the main idea of this work is to consider the following structure:
\begin{eqnarray}\label{key_eq}
\left[ \begin{array}{c}
f_1(h) \\ 
f_2(h)
\end{array} \right] \sim \mathcal{GP}\left( \left[ \begin{array}{c}
0 \\ 
0
\end{array} \right] ,  \left[ \begin{array}{cc}
k_{1}(h,h') & \rho k_{1}(h,h') \\ 
\rho k_{1}(h,h') & \rho^2 k_{1}(h,h') + k_2(h,h')
\end{array}   \right]   \right),
\end{eqnarray}
where
\[
x \longmapsto h := h(x) \longmapsto \left[ \begin{array}{c}
f_1(h(x)) \\ 
f_2(h(x))
\end{array} \right].
\]
The high fidelity system is modeled by $f_2(h(x))$ and the low fidelity one by $f_1(h(x))$. We use $\mathcal{GP}$ to denote a Gaussian Process. This approach can use any deterministic parametric data transformation $h(x)$. However, we focus on multi-layer neural networks
\[
h(x) := (h^L \circ \ldots \circ h^1) (x),
\]
where each layer of the network performs the transformation
\[
h^\ell(z) = \sigma^\ell(w^\ell z + b^\ell),
\]
with $\sigma^\ell$ being the transfer function, $w^\ell$ the weights, and $b^\ell$ the bias of the layer. We use $\theta_h:= [w^1,b^1,\ldots,w^L,b^L]$ to denote the parameters of the neural network. Moreover, $\theta_1$ and $\theta_2$ denote the hyper-parameters of the covariance functions $k_1$ and $k_2$, respectively. The parameters of the model are therefore given by
\[
\theta := [\rho,\theta_1,\theta_2,\theta_h].
\]
It should be noted that the AR(1) Co-kriging model of \cite{kennedy2000predicting} is a special case of our model in the sense that for AR(1) Co-kriging $h = h(x)=x$.

\subsection{AR(1) Co-kriging} In \cite{kennedy2000predicting}, the authors consider the following autoregressive model
\[
f_2(x) = \rho f_1(x) + \delta_2(x),
\]
where $\delta_2(x)$ and $f_1(x)$ are two independent Gaussian Processes with
\[
\delta_2(x) \sim \mathcal{GP}(0,k_2(x,x')),
\]
and
\[
f_1(x) \sim \mathcal{GP}(0,k_1(x,x')).
\]
Therefore,
\[
f_2(x) \sim \mathcal{GP}(0,\rho^2k_1(x,x') + k_2(x,x')),
\]
and
\begin{eqnarray}\label{eq1}
\left[ \begin{array}{c}
f_1(x) \\ 
f_2(x)
\end{array} \right] \sim \mathcal{GP}\left( \left[ \begin{array}{c}
0 \\ 
0
\end{array} \right] ,  \left[ \begin{array}{cc}
k_{1}(x,x') & \rho k_{1}(x,x') \\ 
\rho k_{1}(x,x') & \rho^2 k_{1}(x,x') + k_2(x,x')
\end{array}   \right]   \right),
\end{eqnarray}
which is a special case of (\ref{key_eq}) with $h = h(x) = x$. The importance of $\rho$ is evident from (\ref{eq1}). If $\rho = 0$, the high fidelity and low fidelity models are fully decoupled and by combining there will be no improvements of the prediction.

\section{Prediction} The Deep Multi-fidelity Gaussian Process structure (\ref{key_eq}) can be equivalently written in the following compact form of a multivariate Gaussian Process
\begin{eqnarray}\label{eq}
\left[ \begin{array}{c}
f_1(h) \\ 
f_2(h)
\end{array} \right] \sim \mathcal{GP}\left( \left[ \begin{array}{c}
0 \\ 
0
\end{array} \right] ,  \left[ \begin{array}{cc}
k_{11}(h,h') & k_{12}(h,h') \\ 
k_{21}(h,h') & k_{22}(h,h')
\end{array}   \right]   \right)
\end{eqnarray}
with $k_{11} \equiv k_1, k_{12} \equiv k_{21} \equiv \rho k_1$, and $k_{22} \equiv \rho^2 k_1 + k_2$. This can be used to obtain the predictive distribution
\[
p\left(f_2(h(x_*))|x_*,\bm{x_1},\bm{f}_1,\bm{x_2},\bm{f}_2\right)
\]
of the surrogate model for the high fidelity system at a new test point $x_*$ (see equation \eqref{pred}). Note that the terms $k_{12}(h(x),h(x'))$ and $k_{21}(h(x),h(x'))$ model the correlation between the high-fidelity and the low-fidelity data and therefore are of paramount importance. The key role played by $\rho$ is already well-known in the literature \cite{kennedy2000predicting}. Along the same lines one can easily observe the effectiveness of learning the transformation function $h(x)$ jointly from the low fidelity and high fidelity data.\\

\noindent We obtain the following joint density:
\[
\left[\begin{array}{c}
f_2(h(x_*)) \\ 
\bm{f}_1 \\ 
\bm{f}_2
\end{array}  \right] \sim \mathcal{N}\left( \left[\begin{array}{c}
0 \\ 
\bm{0} \\ 
\bm{0}
\end{array}  \right], \left[ \begin{array}{ccc}
k_{22}(h_*,h_*) & k_{21}(h_*,\bm{h}_1) & k_{22}(h_*,\bm{h}_2) \\ 
k_{12}(\bm{h}_1,h_*) & k_{11}(\bm{h}_1,\bm{h}_1) & k_{12}(\bm{h}_1,\bm{h}_2) \\ 
k_{22}(\bm{h}_2,h_*) & k_{21}(\bm{h}_2,\bm{h}_1) & k_{22}(\bm{h}_2,\bm{h}_2)
\end{array}    \right]   \right),
\]
where $h_* = h(x_*), \bm{h}_1 = h(\bm{x}_1)$, and $\bm{h}_2 = h(\bm{x}_2)$. From this, we conclude that
\begin{equation}\label{pred}
p\left(f_2(h(x_*))|x_*,\bm{x_1},\bm{f}_1,\bm{x_2},\bm{f}_2\right) = \mathcal{N}\left(K_* K^{-1} \bm{f}, k_{22}(h_*,h_*) - K_* K^{-1} K_*^T\right),
\end{equation}
where
\begin{eqnarray}
&&\bm{f} := \left[\begin{array}{c}
\bm{f}_1 \\ 
\bm{f}_2
\end{array}  \right],\\
&&K_* := \left[\begin{array}{cc}
k_{21}(h_*,\bm{h}_1) & k_{22}(h_*,\bm{h}_2)
\end{array}  \right],\\
&&K := \left[\begin{array}{cc}
k_{11}(\bm{h}_1,\bm{h}_1) & k_{12}(\bm{h}_1,\bm{h}_2) \\ 
k_{21}(\bm{h}_2,\bm{h}_1) & k_{22}(\bm{h}_2,\bm{h}_2)
\end{array}    \right].
\end{eqnarray}

\section{Training} The Negative Marginal Log Likelihood $\mathcal{L}(\theta) := -\log p\left( \bm{f} | \bm{x}\right)$ is given by
\begin{eqnarray}\label{Likelihood}
\mathcal{L}(\theta) = \frac12 \bm{f}^T K^{-1}\bm{f} + \frac12 \log \left| K  \right| + \frac{n_1 + n_2}{2}\log 2\pi,
\end{eqnarray}
where
\[
\bm{x} := \left[\begin{array}{c}
\bm{x}_1 \\ 
\bm{x}_2
\end{array}   \right].
\]
The Negative Marginal Log Likelihood along with its Gradient can be used to estimate the parameters $\theta$. Finding the gradient $\frac{\partial\mathcal{L}(\theta)}{\partial \theta}$ is discussed in the following. First observe that
\[
\frac{\partial\mathcal{L}(\theta)}{\partial K} = -\frac{1}{2} K^{-1}\bm{f}\bm{f}^TK^{-1} + \frac12 K^{-1}.
\]
Therefore,
\begin{eqnarray}
\frac{\partial\mathcal{L}(\theta)}{\partial \rho} = \frac{\partial\mathcal{L}(\theta)}{\partial K} \frac{\partial K}{\partial \rho}\\
\frac{\partial\mathcal{L}(\theta)}{\partial \theta_1} = \frac{\partial\mathcal{L}(\theta)}{\partial K} \frac{\partial K}{\partial \theta_1}\nonumber\\
\frac{\partial\mathcal{L}(\theta)}{\partial \theta_2} = \frac{\partial\mathcal{L}(\theta)}{\partial K} \frac{\partial K}{\partial \theta_2}\nonumber,
\end{eqnarray}
and
\begin{eqnarray}
\frac{\partial\mathcal{L}(\theta)}{\partial \theta_h} = \frac{\partial\mathcal{L}(\theta)}{\partial K} \frac{\partial K}{\partial \bm{h}}\frac{\partial \bm{h}}{\partial \theta_h},
\end{eqnarray}
where $\bm{h} = h(\bm{x})$. We use backpropagation to find $\frac{\partial \bm{h}}{\partial \theta_h}$. Backpropagation is a popular method of training artificial neural networks. With this method one can calculate the gradients of $\bm{h}$ with respect to all the parameters $\theta_h$ in the network.

\section{Summary of the Algorithm} The following summarizes our Deep Multi-fidelity GP algorithm.\\
\begin{itemize}
\item First, we employ the Negative Marginal Log Likelihood (see eq. \ref{Likelihood}) to train the parameters and hyper-parameters of the model using the low and high-fidelity data $\{\bm{x},\bm{f}\}$. We are therefore jointly training the neural network $h(x)$ and the kernels $k_1$ and $k_2$ introduced in eq. \ref{key_eq}.\\
\item Then, use eq. \ref{pred} to predict the the output of the high-fidelity function at a new test point $x^*$.
\end{itemize}

\section{Numerical Experiments} To demonstrate the effectiveness of our proposed method, we apply our Deep Multi-fidelity Gaussian Processes algorithm to the following challenging benchmark problems.

\subsection{Step Function} 
\begin{figure}[!ht]
\centering
\includegraphics[scale=.5]{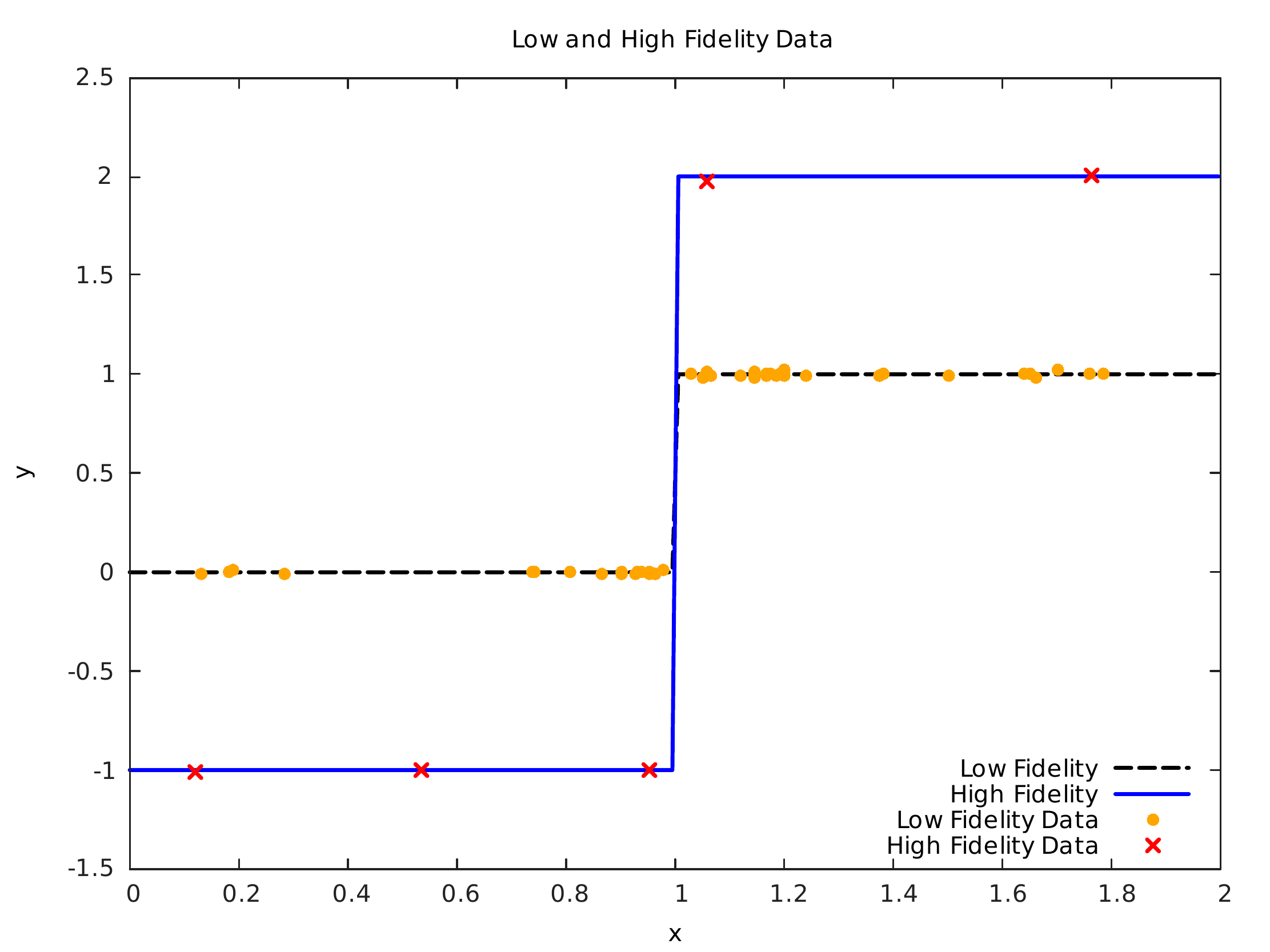}
\caption{Low-fidelity and High-fidelity dataset $\left\{ \{\bm{x}_1, \bm{y}_1\}, \{\bm{x}_2,\bm{y}_2\}\right\}$}\label{StepFunctionDataSet}
\end{figure}

The high fidelity data is generated by the following step function
\[
y_2 = f_2(x) + \epsilon_2, ~~~~~\epsilon_2\sim \mathcal{N}(0,0.01^2)
\]
where
\[
f_2(x) = \left\{\begin{array}{cc}
-1 & \text{if~~} 0 \leq x \leq 1 \\ 
2 & \text{if~~} 1 < x \leq 2
\end{array},  \right.
\]
and the low fidelity data are generated by
\[
y_1 = f_1(x) + \epsilon_1, ~~~~~\epsilon_1\sim \mathcal{N}(0,0.01^2)
\]
where
\[
f_1(x) = \left\{\begin{array}{cc}
0 & \text{if~~} 0 \leq x \leq 1 \\ 
1 & \text{if~~} 1 < x \leq 2
\end{array}.  \right.
\]
In order to generate the training data, we pick $50+100+50$ uniformly distributed random points from the interval $[0,2] = [0,0.8] \cup [0.8,1.2] \cup [1.2,2]$. Out of these $200$ points, $n_1 = 45$ are chosen at random to constitute $\bm{x}_1$ and $n_2 = 5$ are picked at random to create $\bm{x}_2$. We therefore obtain the dataset $\left\{ \{\bm{x}_1, \bm{y}_1\}, \{\bm{x}_2,\bm{y}_2\}\right\}$. This dataset is depicted in figure \ref{StepFunctionDataSet}.\\

\noindent We use a multi-layer neural network of $[3-2]$ neurons. This means that $h:\mathbb{R} \rightarrow \mathbb{R}^2$ is given by
\[
h(x) = h^2(h^1(x)).
\]
Moreover, $h^1:\mathbb{R}\rightarrow \mathbb{R}^3$ is given by $h^1(x) = \sigma(w^1 x + b^1)$ with $\sigma(z) = 1/(1 + e^{-z})$ being the Sigmoid function. Furthermore, $h^2:\mathbb{R}^3 \rightarrow \mathbb{R}^2$ is given by $h^2(z) = w^2 z + b^2$.\\

\begin{figure}[!ht]
\centering
\includegraphics[scale=.5]{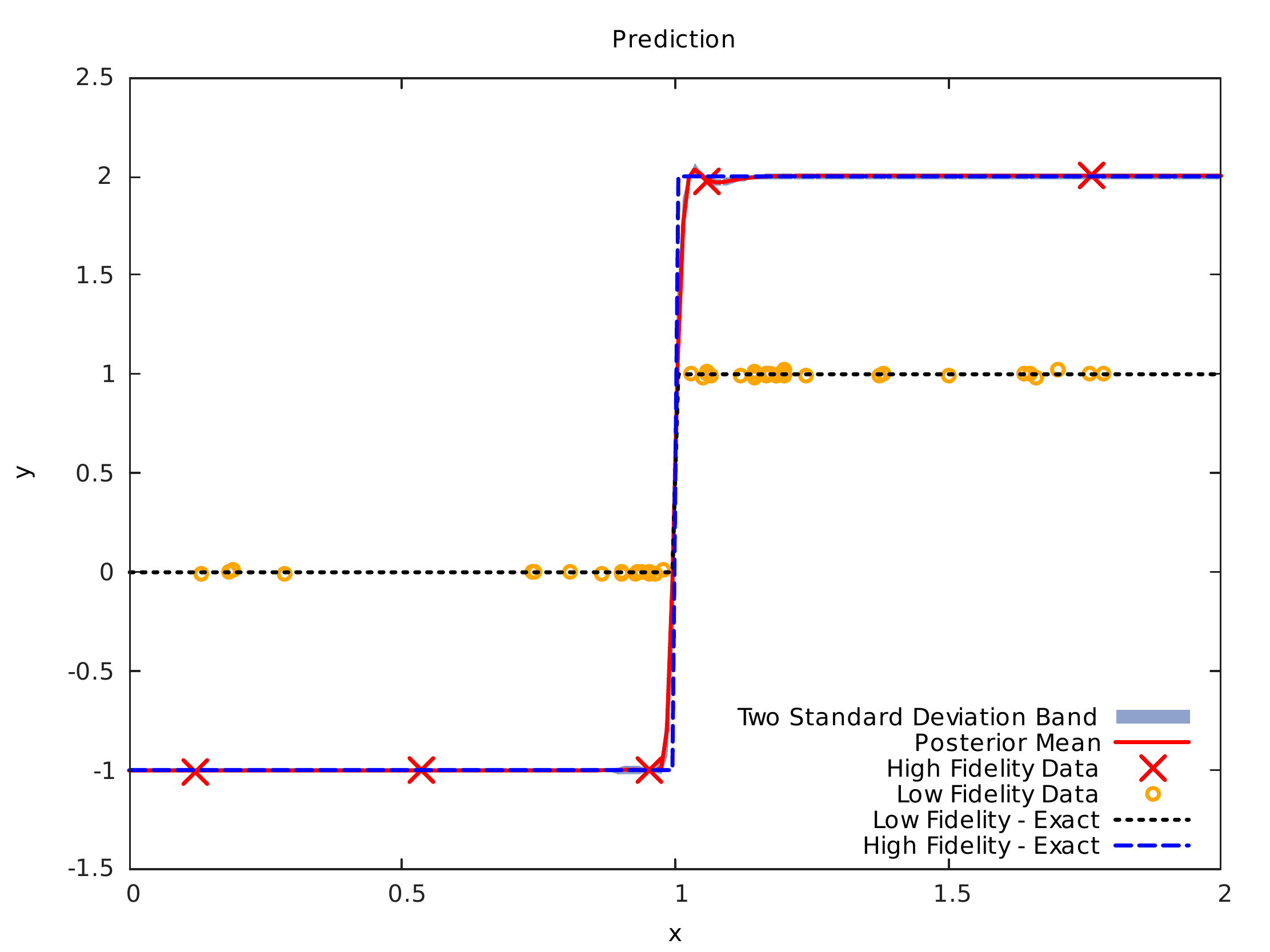}
\caption{Deep Multi-fidelity Gaussian Processes predicive mean and two standard deviations}\label{StepFunctionPrediction}
\end{figure}

\noindent As for the kernels $k_1$ and $k_2$, we use the squared exponential covariance functions with Automatic Relevance Determination (ARD) (see \cite{williams2006gaussian}) of the form
\[
k(x,x') = \sigma_f^2 \exp\left(-\frac12\sum_{d=1}^D \left(\frac{x_d - x_d'}{\ell_d}\right)^2\right).
\]
The predictive mean and two standard deviation bounds for our Deep Multi-fidelity Gaussian Processes method is depicted in figure \ref{StepFunctionPrediction}.\\

\begin{figure}[!ht]
\centering
\includegraphics[scale=.5]{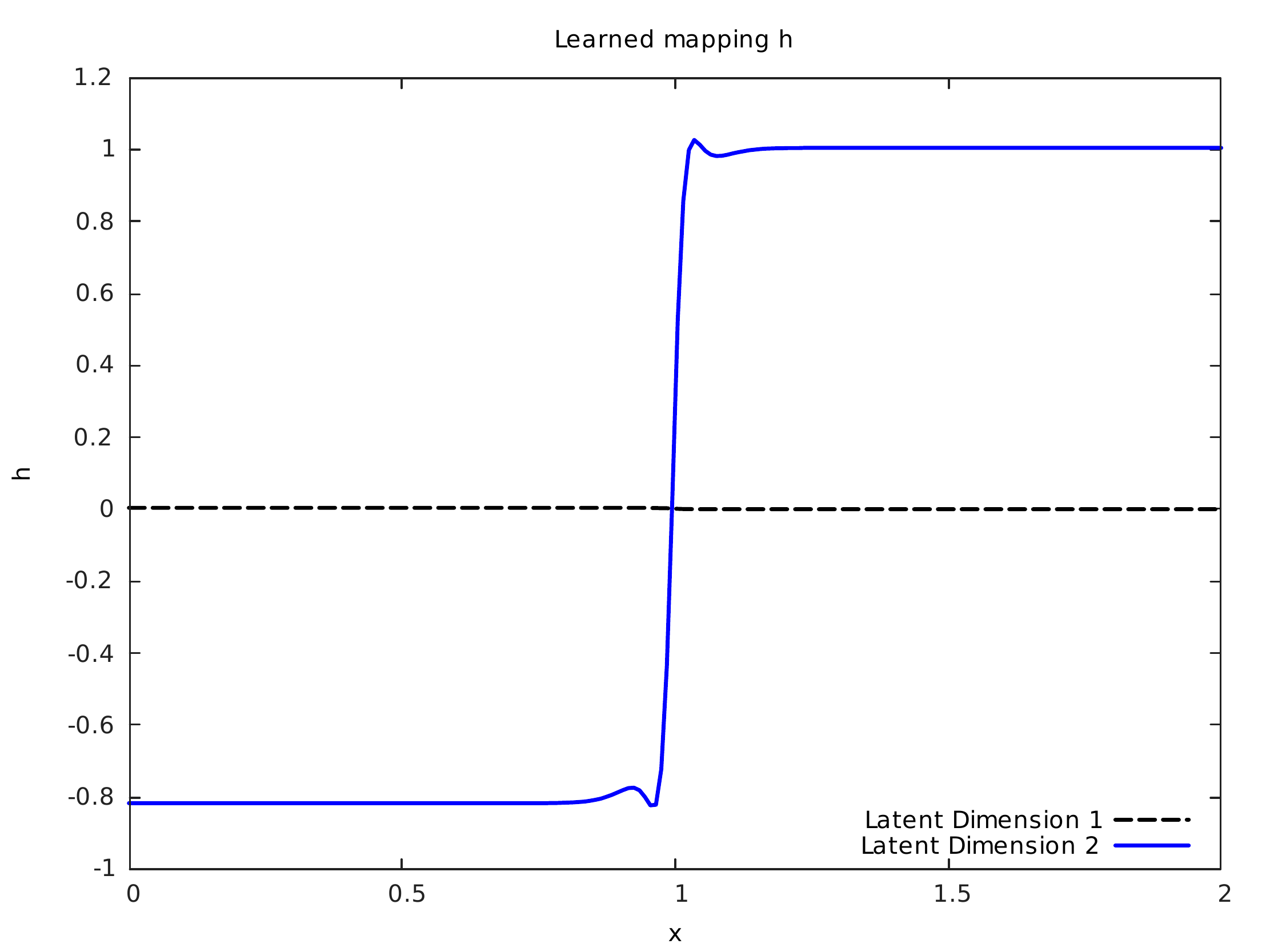}
\caption{The 2D feature space discovered by the nonlinear mapping $h$. Notice that $h:\mathbb{R}\rightarrow\mathbb{R}^2$. Latent dimensions 1 and 2 correspond to the first and second dimensions of $h(\mathbb{R})$.}\label{StepFunctionLearnedH}
\end{figure}

\noindent The 2D feature space discovered by the nonlinear mapping $h$ is depicted in figure \ref{StepFunctionLearnedH}. Recall that, for this example, we have $h:\mathbb{R} \rightarrow \mathbb{R}^2$.\\

\begin{figure}[!ht]
\centering
\includegraphics[scale=.5]{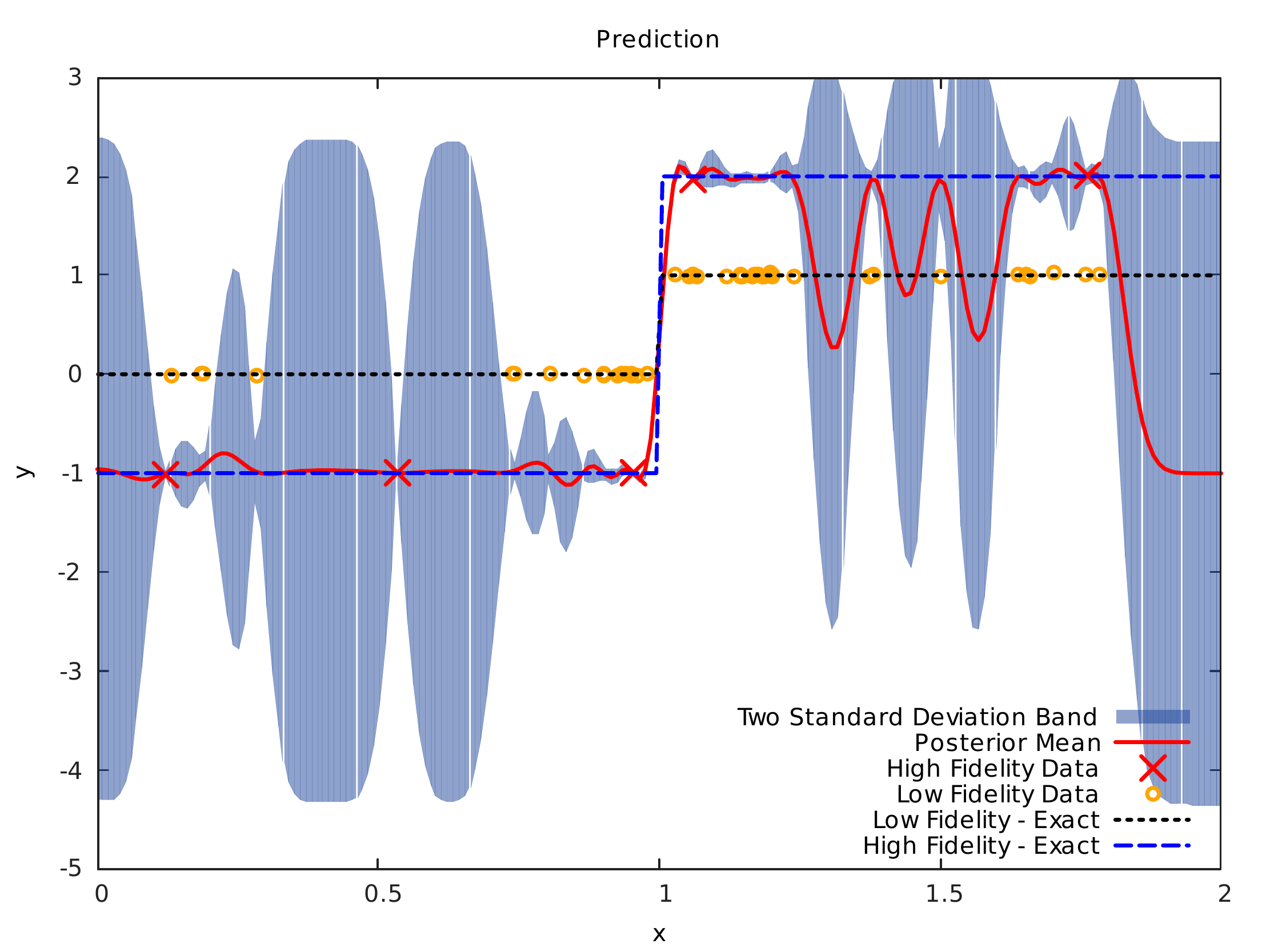}
\caption{AR(1) Co-kriging predicive mean and two standard deviations}\label{StepFunctionPredictionAR1}
\end{figure}

\noindent The discontinuity of the model is captured by the non-linear mapping $h$. Therefore, the mapping from the feature space to outputs is smooth and can be easily handled by a regular AR(1) Co-kriging model. In order to see the importance of the mapping $h$, let us compare our method with AR(1) Co-kriging. This is depicted in figure \ref{StepFunctionPredictionAR1}.\\

\subsection{Forrester Function \cite{forrester2007multi} with Jump} 
\begin{figure}[!ht]
\centering
\includegraphics[scale=.5]{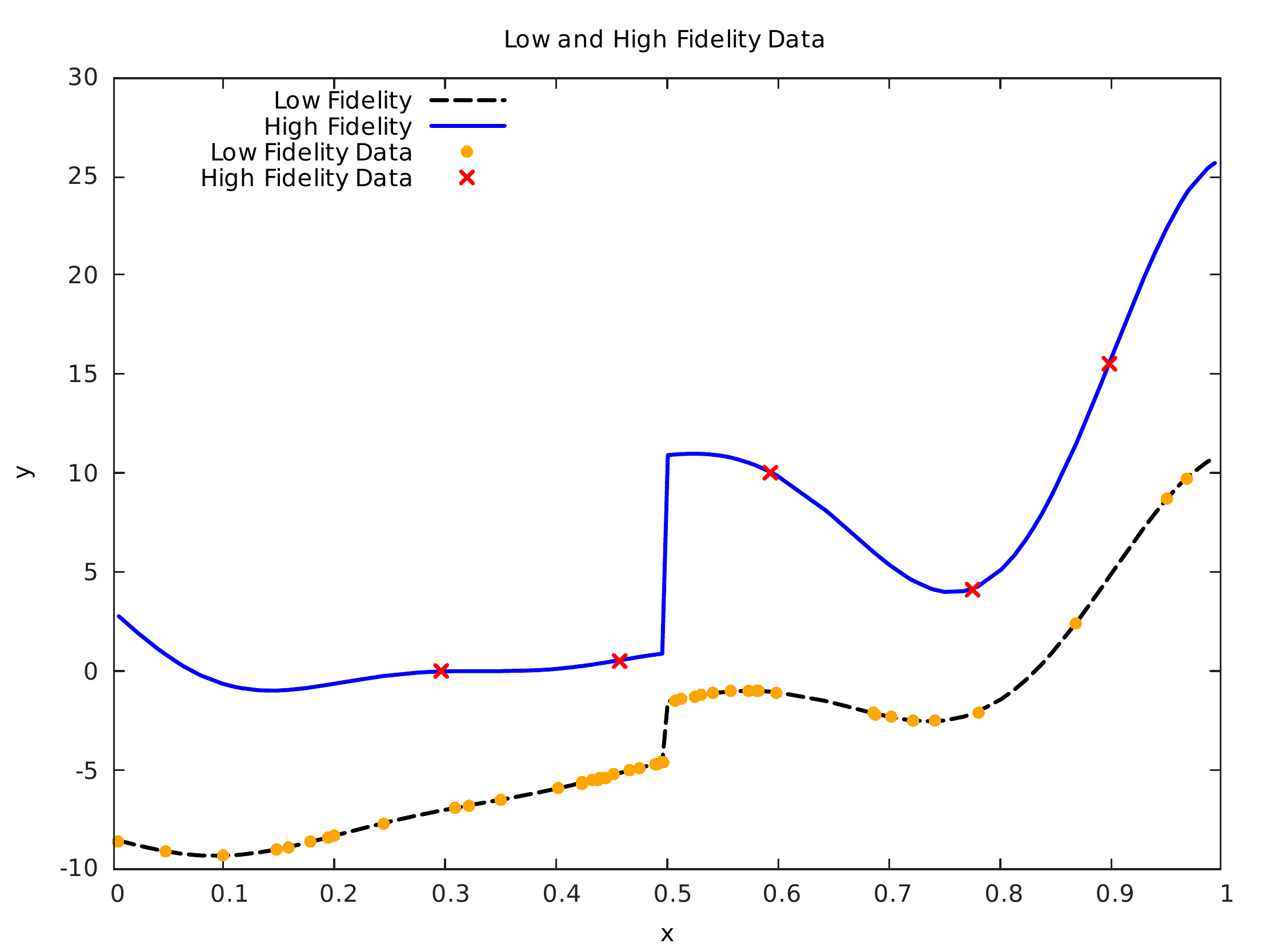}
\caption{Low-fidelity and High-fidelity dataset $\left\{ \{\bm{x}_1, \bm{f}_1\}, \{\bm{x}_2,\bm{f}_2\}\right\}$}\label{ForresterWithJumpDataSet}
\end{figure}

The low fidelity data are generated by
\[
f_1(x) = \left\{\begin{array}{cc}
0.5(6x - 2)^2\sin(12x-4) + 10(x-0.5) - 5 & 0 \leq x \leq 0.5 \\ 
3 + 0.5(6x - 2)^2\sin(12x-4) + 10(x-0.5) - 5 & 0.5 \leq x \leq 1
\end{array}  \right.,
\]
and the high fidelity data are generated by
\[
f_1(x) = \left\{\begin{array}{cc}
2f_1(x) - 20x + 20 & 0 \leq x \leq 0.5 \\ 
4 + 2f_1(x) - 20x + 20 & 0.5 \leq x \leq 1
\end{array}  \right..
\]
In order to generate the training data, we pick $50+100+50$ uniformly distributed random points from the interval $[0,1] = [0,0.4] \cup [0.4,0.6] \cup [0.6,1]$. Out of these $200$ points, $n_1 = 50$ are chosen at random to constitute $\bm{x}_1$ and $n_2 = 5$ are picked at random to create $\bm{x}_2$. We therefore obtain the dataset $\left\{ \{\bm{x}_1, \bm{f}_1\}, \{\bm{x}_2,\bm{f}_2\}\right\}$. This dataset is depicted in figure \ref{ForresterWithJumpDataSet}.\\

\begin{figure}[!ht]
\centering
\includegraphics[scale=.5]{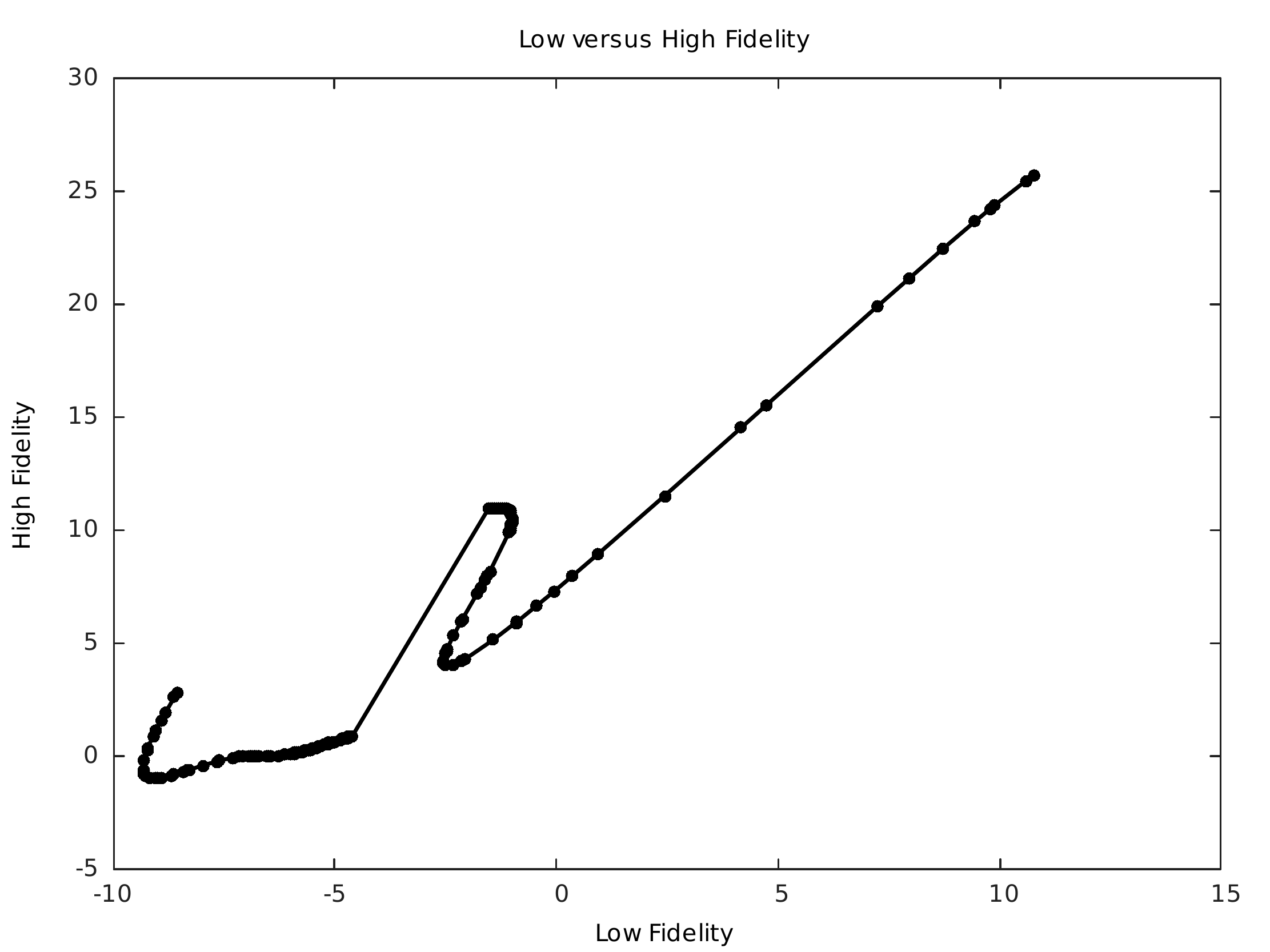}
\caption{Relation between the Low-fidelity and High-fidelity data generating processes.}\label{ForresterWithJumpRelation}
\end{figure}

Figure \ref{ForresterWithJumpRelation} depicts the relation between the low fidelity and the high fidelity data generating processes. One should notice the discontinuous and non-functional form of this relation.

\begin{figure}[!ht]
\centering
\includegraphics[scale=.5]{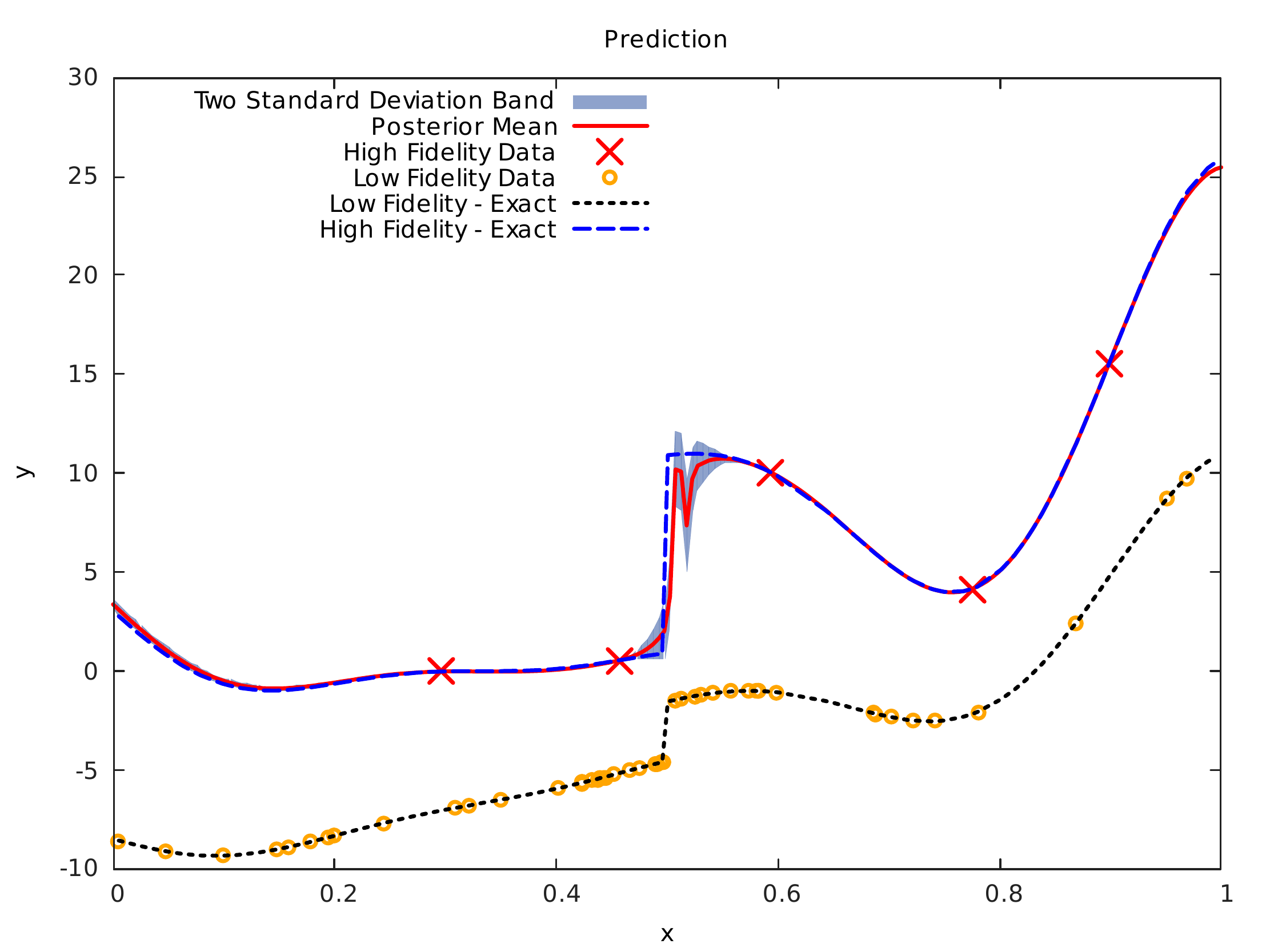}
\caption{Deep Multi-fidelity Gaussian Processes predicive mean and two standard deviations}\label{ForresterWithJumpPrediction}
\end{figure}

\noindent Our choice of the neural network and covariance functions is as before. The predictive mean and two standard deviation bounds for our Deep Multi-fidelity Gaussian Processes method is depicted in figure \ref{ForresterWithJumpPrediction}.\\

\begin{figure}[!ht]
\centering
\includegraphics[scale=.5]{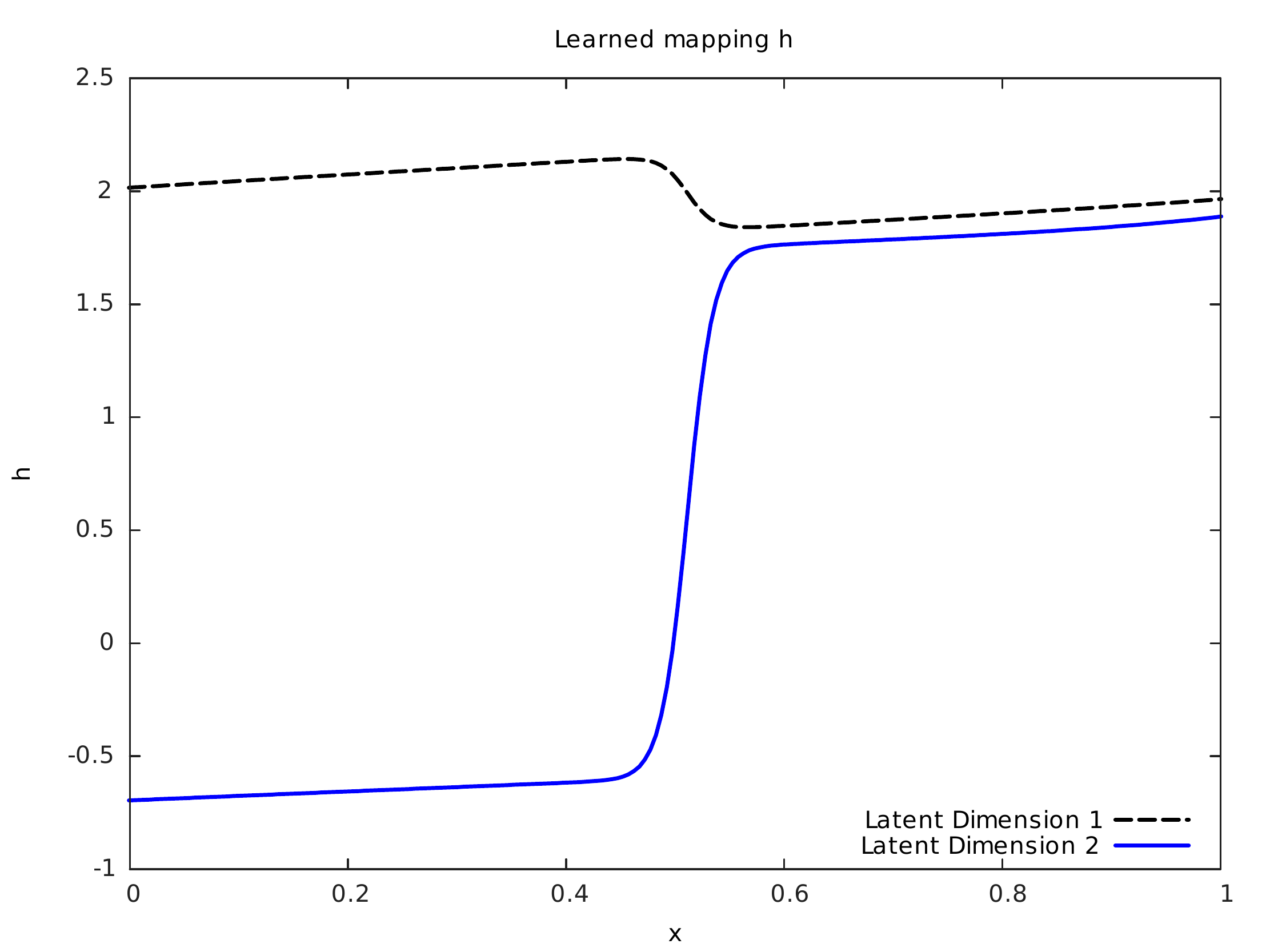}
\caption{The 2D feature space discovered by the nonlinear mapping $h$. Notice that $h:\mathbb{R}\rightarrow\mathbb{R}^2$. Latent dimensions 1 and 2 correspond to the first and second dimensions of $h(\mathbb{R})$.}\label{ForresterWithJumpLearnedH}
\end{figure}

\noindent The 2D feature space discovered by the nonlinear mapping $h$ is depicted in figure \ref{ForresterWithJumpLearnedH}.\\

\begin{figure}[!ht]
\centering
\includegraphics[scale=.5]{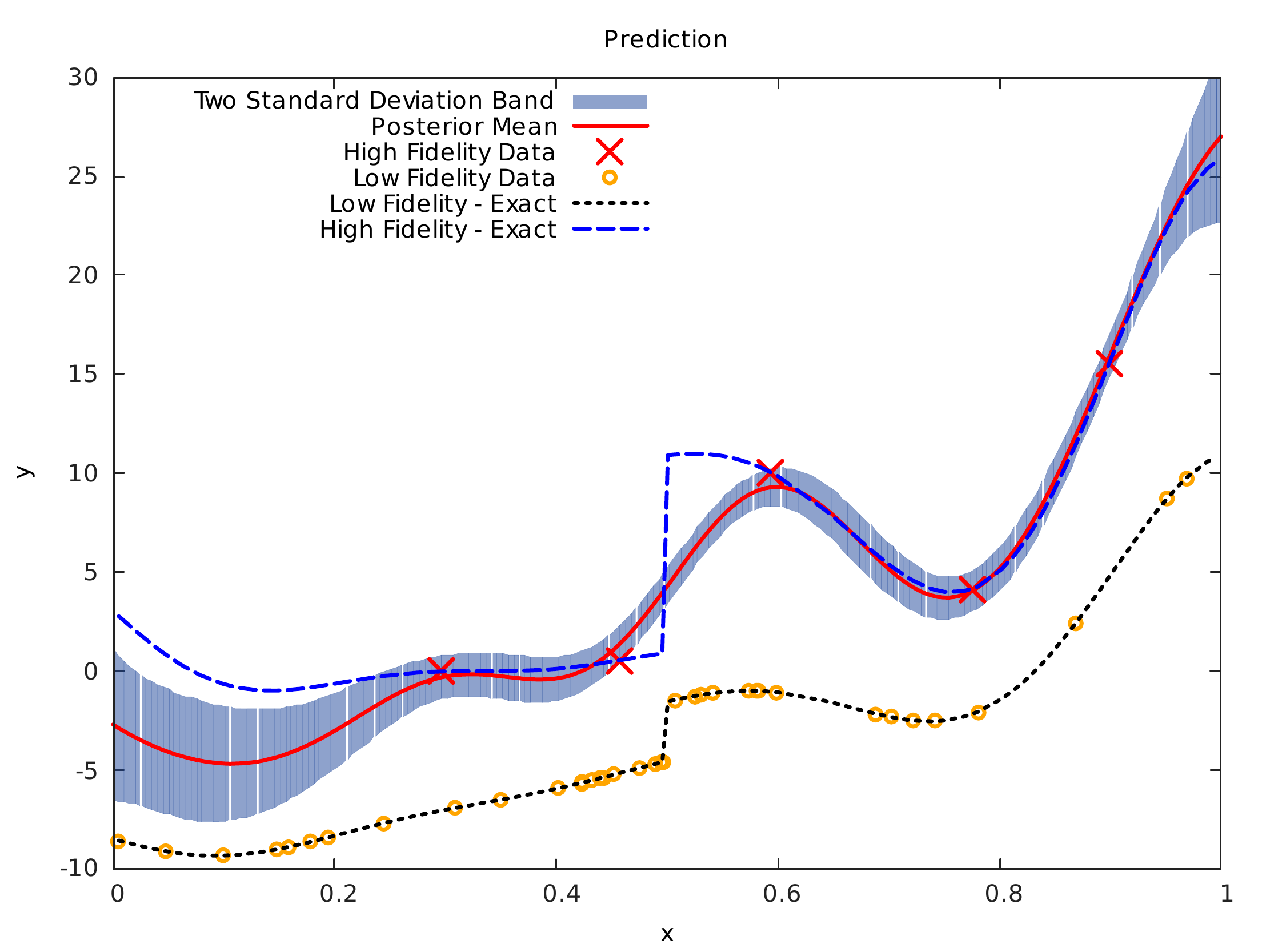}
\caption{AR(1) Co-kriging predicive mean and two standard deviations}\label{ForresterWithJumpPredictionAR1}
\end{figure}

\noindent Once again, the discontinuity of the model is captured by the non-linear mapping $h$. In order to see the importance of the mapping $h$, let us compare our method with AR(1) Co-kriging. This is depicted in figure \ref{ForresterWithJumpPredictionAR1}.\\

\subsection{A Sample Function} 
\begin{figure}[!ht]
\centering
\includegraphics[scale=.5]{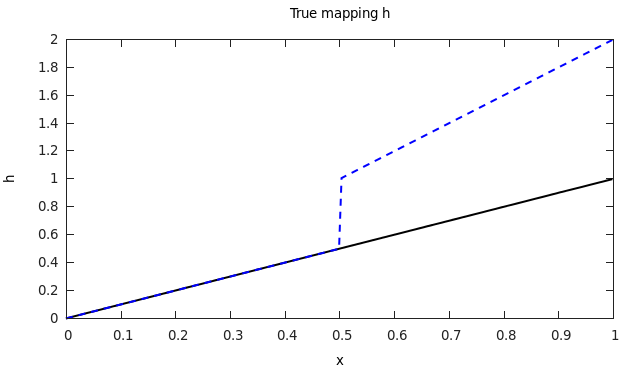}
\caption{The true mapping $h:\mathbb{R}\rightarrow \mathbb{R}^2$.}\label{SampleTrueH}
\end{figure}

The main objective of this section is to demonstrate the types of cross-correlation structures that our framework is capable of handling. In the following, let the true mapping $h:\mathbb{R} \rightarrow \mathbb{R}^2$ be given by
\[
h(x) = \left\{\begin{array}{cc}
\left[\begin{array}{c}
x \\ 
x
\end{array} \right], & 0 \leq x \leq 0.5 \\ 
\\
\left[\begin{array}{c}
x \\ 
2x
\end{array} \right], & 0.5 \leq x \leq 1
\end{array}  \right..
\]
This is plotted in figure \ref{SampleTrueH}.\\

\begin{figure}[!ht]
\centering
\includegraphics[scale=.5]{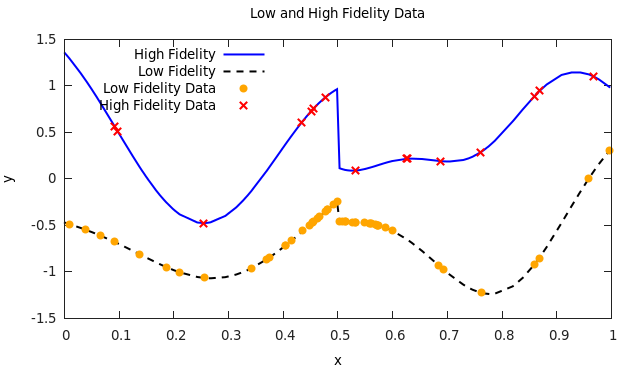}
\caption{Low-fidelity and High-fidelity dataset $\left\{ \{\bm{x}_1, \bm{f}_1\}, \{\bm{x}_2,\bm{f}_2\}\right\}$}\label{SampleDataSet}
\end{figure}

\noindent Given $\rho=1$, we generate a sample of the joint prior distribution \ref{key_eq}. This gives us two sample functions $f_1(x)$ and $f_2(x)$, where $f_2(x)$ is the high-fidelity one. In order to generate the training data, we pick $50+100+50$ uniformly distributed random points from the interval $[0,1] = [0,0.4] \cup [0.4,0.6] \cup [0.6,1]$. Out of these $200$ points, $n_1 = 50$ are chosen at random to constitute $\bm{x}_1$ and $n_2 = 15$ are picked at random to create $\bm{x}_2$. We therefore obtain the dataset $\left\{ \{\bm{x}_1, \bm{f}_1\}, \{\bm{x}_2,\bm{f}_2\}\right\}$. This dataset is depicted in figure \ref{SampleDataSet}.\\

\begin{figure}[!ht]
\centering
\includegraphics[scale=.5]{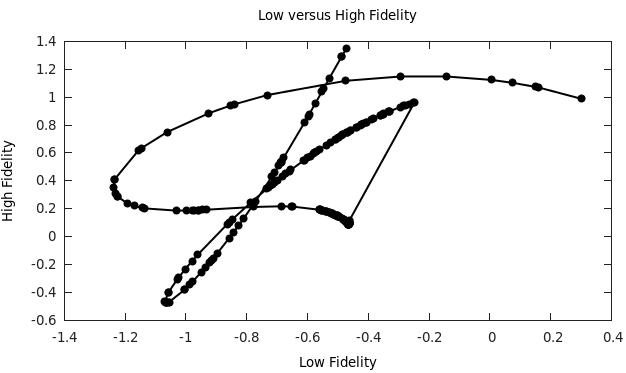}
\caption{Relation between the Low-fidelity and High-fidelity data generating processes.}\label{SampleRelation}
\end{figure}

Figure \ref{SampleRelation} depicts the relation between the low fidelity and the high fidelity data generating processes. One should notice the discontinuous and non-functional form of this relation.\\

\begin{figure}[!ht]
\centering
\includegraphics[scale=.435]{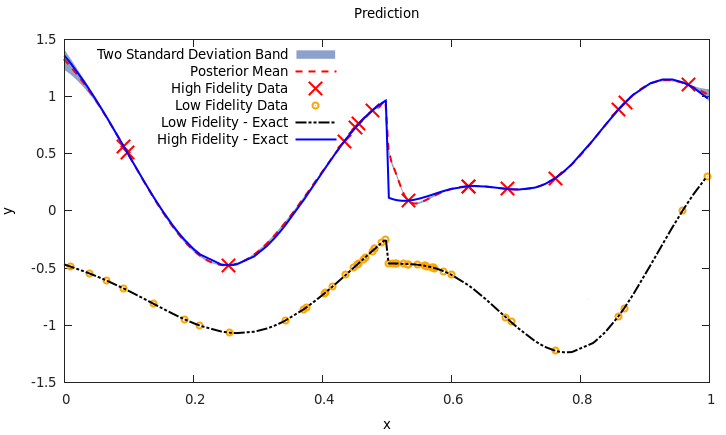}
\caption{Deep Multi-fidelity Gaussian Processes predicive mean and two standard deviations}\label{SamplePrediction}
\end{figure}

\noindent Our choice of the neural network and covariance functions is as before. The predictive mean and two standard deviation bounds for our Deep Multi-fidelity Gaussian Processes method is depicted in figure \ref{SamplePrediction}.\\

\begin{figure}[!ht]
\centering
\includegraphics[scale=.5]{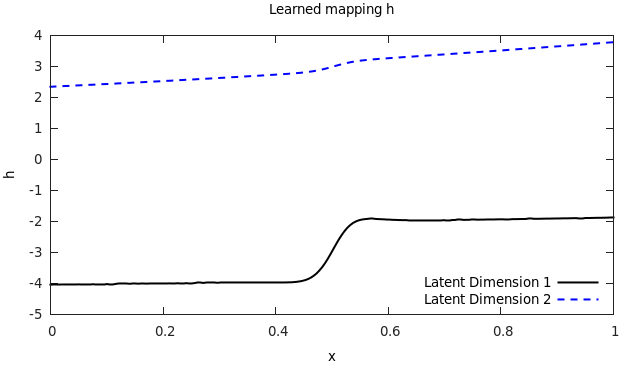}
\caption{The 2D feature space discovered by the nonlinear mapping $h$. Notice that $h:\mathbb{R}\rightarrow\mathbb{R}^2$. Latent dimensions 1 and 2 correspond to the first and second dimensions of $h(\mathbb{R})$.}\label{SampleLearnedH}
\end{figure}

\noindent The 2D feature space discovered by the nonlinear mapping $h$ is depicted in figure \ref{SampleLearnedH}. One should notice the discrepancy between the true mapping $h$ and the one learned by our algorithm. This discrepancy reflects the fact that the mapping from $x$ to the feature space is not necessarily unique.\\

\begin{figure}[!ht]
\centering
\includegraphics[scale=.43]{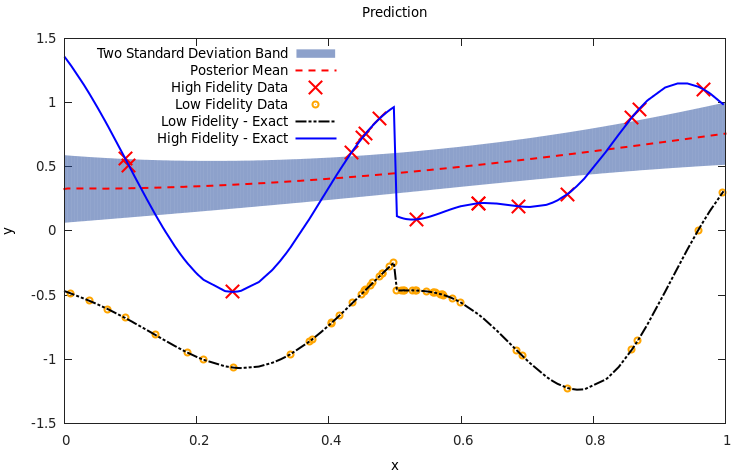}
\caption{AR(1) Co-kriging predicive mean and two standard deviations}\label{SamplePredictionAR1}
\end{figure}

\noindent Once again, the discontinuity of the model is captured by the non-linear mapping $h$. In order to see the importance of the mapping $h$, let us compare our method with AR(1) Co-kriging. This is depicted in figure \ref{SamplePredictionAR1}.\\

\section{Conclusion} We devised a surrogate model that is capable of capturing general discontinuous correlation structures between the low- and high-fidelity data generating processes. The model's efficiency in handling discontinuities was demonstrated using benchmark problems. Essentially, the discontinuity is captured by the neural network. The abundance of low-fidelity data allows us to train the network accurately. We therefore need very few observations of the high-fidelity data generating process.\\

A major drawback of our method could be its overconfidence which stems from the fact that, unlike Gaussian Processes, neural networks are not capable of modeling uncertainty. Modeling the data transformation function $h$ as a Gaussian Process, instead of a neural network, might be a more proper way of modeling uncertainty. However, this becomes analytically intractable and more challenging. This could be a promising subject of future research. A good reference in this direction is \cite{damianou2015deep}.

\section*{Acknowledgments} This work was supported by the DARPA project on Scalable Framework for Hierarchical Design and Planning under Uncertainty with Application to Marine Vehicles (N66001-15-2-4055).

\bibliography{mybib}{}
\bibliographystyle{plain}
\end{document}